\documentclass{article}

\usepackage{PRIMEarxiv}

\usepackage[utf8]{inputenc} 
\usepackage[T1]{fontenc}    
\usepackage{hyperref}       
\usepackage{url}            
\usepackage{booktabs}       
\usepackage{amsfonts}       
\usepackage{nicefrac}       
\usepackage{microtype}      
\usepackage{lipsum}
\usepackage{fancyhdr}       
\usepackage{graphicx}       
\graphicspath{{media/}}     
\usepackage{algorithmic}
\usepackage{float}
\usepackage{bm}
\usepackage{graphicx}
\usepackage{graphics}
\usepackage{caption}
\usepackage{subcaption} 
\usepackage{amsmath}
\usepackage[ruled, vlined]{algorithm2e}

\title{CortiNet: A Physics-Perception Hybrid Cortical-Inspired Dual-Stream Network for Gallbladder Disease Diagnosis from Ultrasound
}

\author{
  Vagish Kumar \\
  Department of Applied Mechanics \\
  Indian Institute of Technology Delhi \\
  \texttt{vagish.kumar@am.iitd.ac.in} \\
   \And
  Souvik Chakraborty \\
  Department of Applied Mechanics \\
  Yardi School of Artificial Intelligence (ScAI) \\
  Indian Institute of Technology Delhi \\
  \texttt{souvik@am.iitd.ac.in} \\
}

\begin{document}
\maketitle

\begin{abstract}
Gallbladder diseases are increasingly prevalent and can lead to severe complications if not diagnosed at an early stage. Ultrasound imaging is the primary diagnostic modality due to its non-invasive nature, affordability, and wide accessibility. However, the low resolution and speckle noise inherent to ultrasound images substantially hinder diagnostic reliability, often prompting the use of deep and computationally heavy convolutional neural networks that are difficult to deploy in routine clinical settings. In this work, we propose CortiNet, a lightweight, cortical-inspired dual-stream neural architecture for automated gallbladder disease diagnosis that integrates physically interpretable multi-scale signal decomposition with perception-driven feature learning. Inspired by \textit{parallel processing pathways in the human visual cortex}, CortiNet explicitly separates low-frequency structural information from high-frequency perceptual details and processes them through specialized encoding streams. By operating directly on structured, frequency-selective representations rather than raw pixel intensities, the architecture embeds strong physics-based inductive bias, enabling efficient feature learning with a significantly reduced parameter footprint. A late-stage \textit{cortical-style} fusion mechanism integrates complementary structural and textural cues while preserving computational efficiency. Additionally, we propose a structure-aware explainability framework wherein gradient-weighted class activation mapping is only applied to the structural branch of the proposed CortiNet architecture. This choice allows the model to only focus on the structural features, making it robust against speckle noise.
We evaluate CortiNet on 10,692 expert-annotated images spanning nine clinically relevant gallbladder disease categories. Experimental results demonstrate that CortiNet achieves high diagnostic accuracy (98.74\%) with only a fraction of the parameters required by conventional deep convolutional models. This combination of accuracy, robustness, and computational efficiency makes CortiNet particularly well suited for deployment in resource-constrained and point-of-care clinical environments. 

\end{abstract}


\section{Introduction}
\label{sec:introduction}
Gallbladder diseases, including cholelithiasis, cholecystitis, gallbladder polyps, and gallbladder carcinoma, represent a growing global health concern with potentially severe clinical consequences if not detected at an early stage \cite{portincasa2016management}. Among these conditions, gallbladder cancer is particularly concerning due to its aggressive progression and lack of specific early symptoms, often resulting in delayed diagnosis and poor prognosis. Epidemiological evidence points to a steady increase in gallbladder disease prevalence driven by lifestyle changes, dietary habits, obesity, and aging populations, while pre-existing conditions such as gallstones significantly elevate the risk of malignant transformation \cite{stinton2012epidemiology, huang2021association}. According to GLOBOCAN 2022, gallbladder cancer alone accounts for approximately 89,055 deaths annually worldwide, with advanced-stage diagnosis associated with a five-year survival rate of only about 5\% \cite{bray2024global, gupta2021locally}. Ultrasonography (USG) is the primary imaging modality for the evaluation of gallbladder abnormalities owing to its non-invasive nature, cost-effectiveness, real-time capability, and widespread availability \cite{avola2021ultrasound}. It plays a central role in detecting gallstones, assessing gallbladder wall thickening, identifying polyps, and providing early indications of malignancy. However, the diagnostic process is challenged by inherent limitations of ultrasound imaging, including speckle noise, low contrast, limited spatial resolution, and operator dependence, which complicate the identification of subtle pathological features and contribute to inter-observer variability \cite{szabo2013diagnostic, lodhi2020accuracy}. These challenges are further compounded by the rapidly increasing volume of imaging studies; recent analyses suggest that, to meet current clinical demands, an average radiologist would need to interpret an image approximately every 3 to 4 seconds, raising concerns about diagnostic consistency, fatigue, and delayed detection \cite{mcdonald2015effects, medicus2025radiologist}.

In response to the increasing diagnostic burden and inherent challenges of ultrasound interpretation, Deep Neural Networks (DNNs), particularly Convolutional Neural Networks (CNNs), have emerged as powerful tools for automated medical image analysis \cite{egger2022medical, fallahpoor2024deep}. CNNs have demonstrated remarkable performance across diverse applications, including skin cancer detection \cite{wang2025self, jojoa2021melanoma, bratchenko2022classification}, lung nodule malignancy prediction \cite{paul2020convolutional, nguyen2023manet}, tuberculosis diagnosis \cite{lakhani2017deep, hu2022high}, brain tumor identification \cite{saleh2020brain, choi2023single, appiah2024brain}, breast cancer detection \cite{gonccalves2022cnn, dai2025multi, aumente2025multi}, and COVID-19 diagnosis \cite{alazab2020covid, gulakala2023rapid, ju2024codenet}. Their prevalence in medical imaging stems from their ability to automatically learn hierarchical and discriminative features, reducing the need for labor-intensive and potentially biased feature engineering. This capability has naturally motivated efforts to apply CNNs to gallbladder disease detection from ultrasound images. However, the increasing reliance on CNNs in clinical decision-making has raised concerns regarding model explainability. These models are perceived as a black-box system. To address this challenge, post-hoc explainability techniques such as Gradient-weighted Class Activation Mapping (Grad-CAM) \cite{selvaraju2017grad} have been widely adopted to provide visual explanations by highlighting image regions that contribute most strongly to a model's predictions \cite{tjoa2020survey}.

Early deep learning (DL) approaches for automated gallbladder (GB) disease diagnosis from ultrasound images primarily focused on demonstrating feasibility in constrained diagnostic settings. Jeong et al. \cite{jeong2020deep} introduced a CNN-based framework for binary classification of benign versus malignant GB polyps using manually selected regions of interest, highlighting the potential of deep models for ultrasound analysis but limiting scalability due to manual intervention and restricted diagnostic scope. Subsequent studies explored automated region emphasis and the use of deeper convolutional backbones, including VGG, Inception, ResNet, and MobileNet architectures, to improve classification performance \cite{obaid2023detection}. While these methods achieved improved accuracy, they largely relied on pixel-domain learning and deep, over-parameterized models, raising concerns regarding computational efficiency, robustness to ultrasound artifacts, and practical deployment in clinical environments. More recent medical artificial intelligence (AI) research has investigated structured representations and advanced architectural designs to better capture the heterogeneous visual characteristics of gallbladder pathologies. GBCNet \cite{basu2022surpassing} incorporated multi-scale feature aggregation and higher-order statistical modeling, along with curriculum learning strategies to mitigate texture bias, demonstrating the benefits of richer representations. RadFormer \cite{basu2023radformer} further integrated global anatomical context with localized pathological cues using attention mechanisms and transformer-based reasoning. Despite their strong diagnostic performance, such architectures introduce substantial model complexity and computational overhead. In contrast, the present work introduces CortiNet, a lightweight cortical-inspired dual-stream architecture that embeds physics-aware multi-scale signal decomposition and perception-driven feature learning as first-class components of an end-to-end framework. By explicitly encoding domain-aware inductive biases and separating structural and perceptual information pathways, CortiNet addresses key challenges in medical AI for ultrasound, namely robustness, efficiency, and deployability, while maintaining high diagnostic accuracy.

Despite these successes, the application of CNNs to ultrasound-based GB disease diagnosis faces several challenges. Medical images are often of relatively low quality, with limited resolution and inherent speckle noise, and annotated datasets for GB abnormalities are considerably smaller than in general computer vision tasks. These factors can degrade model performance, as networks may focus on irrelevant textures rather than clinically meaningful features, such as the shape, boundaries, or morphology of pathological regions \cite{sharma2022deep}. Model explainability is therefore critical in this setting. Visualization techniques help verify whether network attention is focused on clinically relevant gallbladder regions rather than background patterns or imaging artifacts \cite{singh2020explainable}. Furthermore, traditional deep CNNs are computationally demanding, requiring substantial memory and processing resources, which limits their feasibility for real-time deployment in resource-constrained clinical environments. Consequently, while DNNs hold great promise for improving diagnostic consistency and efficiency, their effective translation to GB ultrasound analysis remains challenging, with relatively few studies exploring this specific domain.

The limitations of conventional DL approaches for GB ultrasound analysis motivate the development of architectures that are both computationally efficient and robust to image noise and variability. In particular, designing networks that incorporate domain-specific knowledge, such as the wave-based physics of ultrasound signal formation and the perceptual strategies employed by human experts, can enhance diagnostic reliability while reducing the risk of overfitting to spurious features. Additionally, the need for deployment in resource-constrained clinical environments requires models that maintain high accuracy with a minimal parameter footprint. Inspired by the parallel processing streams of the human visual cortex, dual-stream architectures offer a natural framework for separating complementary information, such as coarse structural patterns and fine-grained details, and integrating them in a coherent manner. By combining physics-aware multi-scale representations with perception-driven feature learning, such architectures have the potential to achieve robust, interpretable, and lightweight diagnostic performance, addressing both the computational and clinical challenges inherent to gallbladder disease detection from ultrasound images. Within this framework, explainability techniques such as Grad-CAM complement the architectural design choices by providing qualitative visual evidence, thereby supporting clinically meaningful decision support alongside quantitative performance. 
This perspective motivates the design of CortiNet, a cortical-inspired dual-stream network that unifies physics-based inductive bias with perceptual feature integration to enable accurate and efficient GB disease diagnosis. The key scientific contributions are as follows:
\begin{itemize}
    \item \textbf{Cortical-Inspired Dual-Stream Architecture}: Introduces CortiNet, a dual-stream network inspired by the parallel processing pathways of the human visual cortex, explicitly separating low-frequency structural information from high-frequency textural details to enhance feature representation.
    \item \textbf{Physics-Perception Hybrid Design}: Integrates wavelet-based multi-scale decomposition with perception-driven encoding, embedding the physics of ultrasound image formation into the network while leveraging complementary perceptual cues.
    \item \textbf{Lightweight and Efficient}: Achieves high diagnostic performance with a minimal parameter footprint, enabling deployment in resource-constrained and point-of-care clinical environments without sacrificing accuracy.
    \item \textbf{Robust Feature Learning}: Improves reliability in noisy and low-contrast ultrasound images by emphasizing clinically relevant features such as GB morphology, boundaries, and fine-grained pathological patterns.
    \item \textbf{Model explainability}:
    Proposes a structure-aware explainability framework that is robust against noise present in USG data. This helps build trust in the underlying model. 
\end{itemize}

The structure of the remaining sections of the paper is as follows. In Section \ref{sec:proposed_approach}, details on the proposed method is presented. Section \ref{sec:results} showcases the various experiments conducted and the performance of the model. Section \ref{sec:discussion} presents key insights into the proposed approach and the results obtained.
Finally, Section \ref{sec:conclusion} provides the concluding remarks and discusses potential future directions for expanding the scope of the work.

\section{CortiNet: A Physics-Perception Hybrid, Cortical-Inspired Dual-Stream Network}\label{sec:proposed_approach}
In this section, we present the proposed neural architecture. We term the proposed architecture CortiNet, short for Cortical-Inspired Dual-Stream Network, to emphasize its design philosophy rooted in both human visual perception and physically grounded signal representations. CortiNet integrates principled multi-scale signal decomposition with neural architectures that mirror parallel processing streams in the visual cortex, enabling robust and interpretable feature learning for noisy medical imaging modalities such as ultrasound. CortiNet is designed as a physics-perception hybrid architecture, where physically interpretable signal decomposition and biologically inspired visual processing are tightly coupled. From a physics standpoint, medical images, particularly ultrasound, are the result of complex wave–tissue interactions, inherently embedding information across multiple spatial frequencies. From a perceptual standpoint, the human visual system efficiently interprets such signals by decomposing them into scale- and orientation-selective components, processed through parallel cortical pathways. By aligning these two principles, CortiNet explicitly separates low-frequency structural information from high-frequency perceptual details and processes them through specialized encoding streams, resulting in improved robustness to noise and enhanced diagnostic representation learning.

\subsection{Perceptually Grounded, Physically Interpretable Decomposition}
The first stage of CortiNet performs a physically meaningful multi-resolution decomposition of the input image, analogous to the frequency-selective receptive fields observed in early visual cortex. Rather than relying solely on learned filters, we incorporate a mathematically grounded wavelet transformation that decomposes the image into complementary components corresponding to different spatial scales and orientations.
For an input image $\bm x\in \mathbb R^{(C\times H\times W)}$, this decomposition yields
\begin{equation}\label{eq:dwt_level1}
    P(x)=\{A_1,D_1^{HL},D_1^{LH},D_1^{HH}\},
\end{equation}
where $A_1$ encodes coarse structural content, while $D_1^{HL}$, $D_1^{LH}$, $D_1^{HH}$ capture orientation-aware high-frequency responses. Implemented using Daubechies basis functions, this decomposition provides compact spatial support, orthogonality, and robustness to ultrasound-specific noise artifacts, thereby embedding physical inductive bias directly into the learning pipeline.
\subsection{Dual Cortical Streams for Structure and Detail}
Inspired by parallel processing pathways in the visual cortex, CortiNet employs two independent encoders with disjoint parameters: (a) \textbf{Structural Stream (Approximation Pathway)}, which learns global anatomical context, spatial continuity, and organ-level morphology and (b) \textbf{Detail Stream (Perceptual Pathway)}, which learns localized textures, edges, and directional variations critical for clinical discrimination. Let $\mathcal S(\cdot ; \bm \theta_S)$ and 
$\mathcal T(\cdot;\bm \theta_T)$ denote the structural and detail encoders, respectively. The corresponding feature representations are
\begin{equation}\label{eq:approx_encoder}
    F_S=\mathcal S(A_1;\bm \theta_S),
\end{equation}
\begin{equation}\label{eq:detail_encoder}
    F_T=\mathcal T([D_1^{HL},D_1^{LH},D_1^{HH}];\bm \theta_T).
\end{equation}
This explicit separation reflects cortical specialization and prevents high-frequency noise from contaminating global structural representations.

\subsection{Cortical Integration via Late Fusion}
\begin{figure}[ht!]
    \centering
    \includegraphics[width=\linewidth]{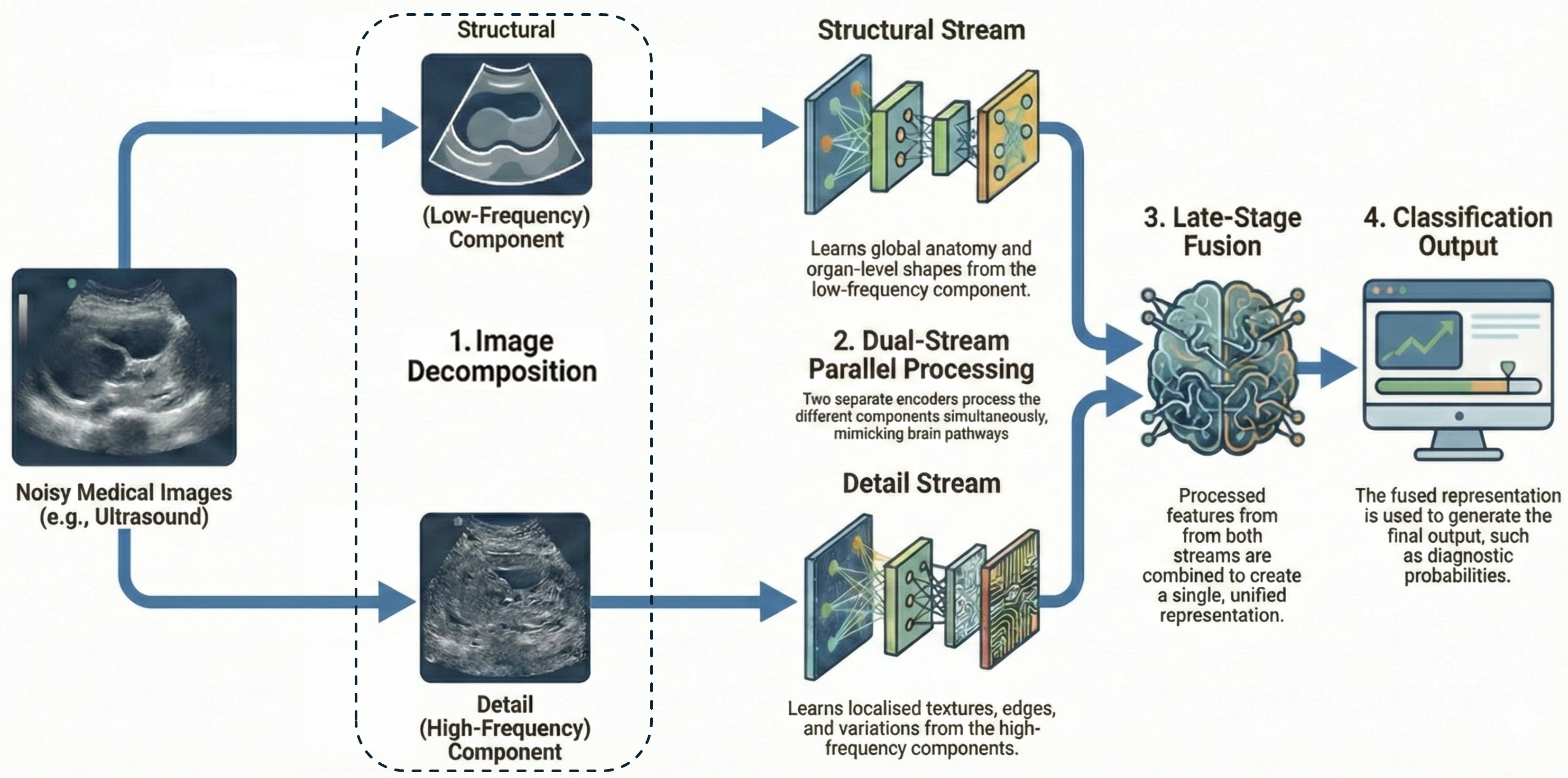}
    \caption{Overview of the proposed CortiNet architecture for ultrasound-based gallbladder disease diagnosis. The framework adopts a cortical-inspired dual-stream design that explicitly separates structural and perceptual information pathways. The input ultrasound image is first transformed into structured, frequency-selective representations using multi-scale signal decomposition, yielding complementary low-frequency components that encode global anatomical structure and high-frequency components that capture fine-grained textural and edge information. These components are processed in parallel by two specialized encoding streams with lightweight convolutional blocks, enabling efficient and targeted feature extraction under ultrasound-specific noise and resolution constraints. To integrate complementary cues, a late-stage cortical-style fusion module combines the learned structural and perceptual representations while preserving computational efficiency and minimizing parameter growth. The fused representation is subsequently passed to a compact classification head that produces multi-class gallbladder disease predictions. By operating on multi-resolution representations and embedding strong domain-aware inductive biases, CortiNet achieves robust diagnostic performance with a significantly reduced parameter footprint, making it suitable for deployment in resource-constrained and point-of-care clinical settings.}
    \label{fig:cortinet}
\end{figure}

In human vision, perceptual integration occurs only after specialized processing in early cortical regions. CortiNet mirrors this principle through late-stage feature fusion. Adaptive global average pooling produces compact, scale-invariant descriptors,
\begin{equation}\label{eq:gap_a,eq:gap_d}
    \mathbf f_S=GAP(F_S),\;\;\; 
    \mathbf f_T=GAP(F_T),
\end{equation}
which are concatenated to form a unified perceptual embedding,
\begin{equation}\label{eq:late_fusion}
    \mathbf f=[\mathbf f_S, \mathbf f_T].
\end{equation}
This fusion mechanism enables joint reasoning over complementary structural and perceptual cues while preserving their semantic distinction. The fused representation is mapped to class probabilities via a fully connected layer with softmax activation, 
\begin{equation}
Z \;=\; \mathbf{f}\,W^\top + b,
\label{eq:fcn}
\end{equation}
\begin{equation}
\hat{\mathbf{y}} = \mathrm{softmax}(Z),
\label{eq:softmax}
\end{equation}
where W and b represent the weights and biases in the fully connected layer. 
The complete parameter set
\begin{equation}
    \bm \theta_{\text{NN}} =\bm \theta_S \cup \bm \theta_T \cup \{\bm W, \bm b\}
\end{equation}
is optimized using the multi-class cross-entropy loss, yielding an end-to-end trainable system grounded in both physical signal characteristics and perceptual processing principles.
A schematic representation of the overall architecture is shown in Fig. \ref{fig:cortinet}. The overall algorithm detailing the training stage is shown in Algorithm \ref{algo:algorithm}.

\begin{algorithm}[ht!]
\SetKwInput{Input}{Input}
\SetKwInput{Output}{Output}
\SetAlgoLined
\LinesNumbered
\Input{A set of images $X=\{x_1, x_2, \ldots, x_N\}$ where $x_n \in \mathbb{R}^{C \times H \times W}$ 
and corresponding labels $C=\{c_1, c_2, \ldots, c_N\}$ where $c_{n} \in \mathbb{Z}^{+}$, and the hyperparameters}
\Output{Optimized network parameters $\theta_{\text{NN}}^\star$}
\BlankLine
\For{$t \leftarrow 1$ \KwTo $T$}{
  \ForEach{mini-batch $\{(x_i,y_i)\}_{i\in\mathcal{B}}$, $|\mathcal{B}|=B$}{
  
    \tcp{Wavelet decomposition}
    Compute DWT of $x_i$: 
    $\mathcal{W}_{\psi}^{(1)}(x_i) = \{A_1, D_1^{\mathrm{HL}}, D_1^{\mathrm{LH}}, D_1^{\mathrm{HH}}\}$; \hfill $\triangleright$ Eq.~\eqref{eq:dwt_level1}\\ 

    \tcp{Dual encoder}
    Pass $A_1$ through aproximation encoder:
    $F_{S} \leftarrow \mathcal{S}(A_1;\theta_{\mathrm{A}})$;  \hfill $\triangleright$ Eq.~\eqref{eq:approx_encoder}\\
    Pass detail coefficients through detail encoder\\
    $F_{T} \leftarrow \mathcal{T}([D_1^{\mathrm{HL}},D_1^{\mathrm{LH}},D_1^{\mathrm{HH}}];\theta_{\mathrm{D}})$;  \hfill $\triangleright$ Eq.~\eqref{eq:detail_encoder}\\

    \tcp{Late fusion}
    Apply adaptive global average pooling\\
    $\mathbf{f}_{S} \leftarrow \mathrm{GAP}(F_{S})$; \quad
    $\mathbf{f}_{T} \leftarrow \mathrm{GAP}(F_{T})$; \hfill $\triangleright$ Eq.~\eqref{eq:gap_a,eq:gap_d}\\
    Concatenate to form the fused representation:
    $\mathbf{f} \leftarrow [\mathbf{f}_{S};\mathbf{f}_{T}]$; \hfill $\triangleright$ Eq.~\eqref{eq:late_fusion}\\

    \tcp{Classification head}
    Map the fused feature to logits:
    $Z \leftarrow \mathbf{f} W^\top + b$; \hfill $\triangleright$ Eq.~\eqref{eq:fcn}\\ 
    Apply softmax to obtain class probabilities:
    $\hat{\mathbf{y}} \leftarrow \mathrm{softmax}(Z)$; \hfill $\triangleright$ Eq.~\eqref{eq:softmax}\\

    \tcp{Loss and parameter update}
    Compute cross-entropy loss: 
    $\ell(\bm \theta_{\text{NN}}) = -\frac{1}{B}\sum_{i\in\mathcal{B}}\sum_{k=1}^{K} \mathbf{1}[y_i=k]\log \hat{y}_{i,k}$; \\
    Compute gradients $\nabla \ell(\bm \theta_{\text{NN}})$; \\
    Update parameters: $\bm{\theta}_{\text{NN}} \leftarrow \mathrm{AdamUpdate}\!\left(\bm{\theta}_{\text{NN}}, \nabla \ell(\bm{\theta}_{\text{NN}})\right)$; \\

  }
}
\caption{Algorithm of the proposed approach}
\label{algo:algorithm}
\end{algorithm}

\subsection{Adaptive Branch Selection in CortiNet for Noise-Aware Inference}
Ultrasound images are inherently affected by speckle noise and acquisition-dependent artifacts, which disproportionately contaminate high-frequency signal components. To explicitly account for this characteristic, CortiNet employs a noise-aware adaptive inference strategy that dynamically determines whether the perceptual (detail) pathway should be activated during inference. The decision is guided by the empirical reliability of the structural (approximation) pathway, which predominantly encodes low-frequency anatomical information and is intrinsically more robust to noise.

Let the structural pathway, leading upto the prediction, be denoted as
\begin{equation}
f_a : \mathcal{X} \rightarrow \Delta^C,
\end{equation}
where $\mathcal{X}$ represents the space of ultrasound images and $\Delta^C$ denotes the $C$-class probability simplex. Given a small calibration dataset
\begin{equation}
\mathcal{D}_{\mathrm{cal}} = \{(x_i, y_i)\}_{i=1}^{N_C},
\end{equation}
drawn from the same distribution as the test data, we evaluate the empirical classification accuracy of the structural pathway as
\begin{equation}
\mathcal{A}_a = \frac{1}{N} \sum_{i=1}^{N_C}
\mathbb{I}
\left[
\arg\max f_a(x_i) = y_i
\right],
\label{eq:calibration_accuracy}
\end{equation}
where $\mathbb{I}(\cdot)$ denotes the indicator function.
This calibration accuracy serves as a distribution-level proxy for structural reliability and effective noise severity. A threshold $\gamma \in (0,1)$ is introduced to control the activation of the perceptual pathway. Specifically, the inference strategy is defined as
\begin{equation}
\text{Activate perceptual pathway} \iff \mathcal{A}_a < \gamma.
\label{eq:branch_decision}
\end{equation}
When $\mathcal{A}_a \ge \gamma$, the structural pathway alone is used for prediction, thereby suppressing noise-dominated high-frequency information and reducing computational overhead.

\begin{algorithm}[ht!]
\caption{Accuracy-Guided Noise-Aware Adaptive Inference in CortiNet}
\label{alg:adaptive_inference}
\SetAlgoLined
\DontPrintSemicolon
\KwIn{Test ultrasound image $x$, calibration dataset $\mathcal{D}_{\mathrm{cal}}$, accuracy threshold $\gamma$}
\KwOut{Predicted gallbladder disease label $\hat{y}$}

\BlankLine
\textbf{Calibration Phase:}\\
Initialize counter $c \leftarrow 0$\;
\ForEach{$(x_i, y_i) \in \mathcal{D}_{\mathrm{cal}}$}{
    Compute $p_a(y \mid x_i) \leftarrow f_a(x_i)$\;
    \If{$\arg\max p_a(y \mid x_i) = y_i$}{
        $c \leftarrow c + 1$\;
    }
}
Compute calibration accuracy $\mathcal{A}_a \leftarrow \frac{c}{|\mathcal{D}_{\mathrm{cal}}|}$\;

\BlankLine
\textbf{Inference Phase:}\\
\eIf{$\mathcal{A}_a \ge \gamma$}{
    \tcp{Structural pathway deemed reliable; perceptual details likely noise-dominated}
    Compute $p_a(y \mid x)$\;
    $\hat{y} \leftarrow \arg\max p_a(y \mid x)$\;
}{
    \tcp{Reduced structural accuracy indicates noise or diagnostic ambiguity}
    Activate perceptual pathway and fusion module\;
    Compute fused prediction $p_f(y \mid x)$\;
    $\hat{y} \leftarrow \arg\max p_f(y \mid x)$\;
}
\Return{$\hat{y}$}
\end{algorithm}

From a statistical learning perspective, high-frequency image representations tend to exhibit lower bias but higher variance, particularly under noisy measurement processes such as ultrasound acquisition. In contrast, low-frequency structural representations introduce greater bias but substantially reduce variance, yielding improved robustness under distributional shifts. Unconditional fusion of both representations can therefore amplify noise-driven features, leading to degraded generalization.
The proposed accuracy-guided adaptive inference strategy explicitly exploits this bias-variance trade-off. The structural pathway acts as a noise-robust estimator, while the perceptual pathway functions as a conditional refinement mechanism that is activated only when supported by distribution-level reliability evidence. Calibration accuracy serves as a principled, data-driven estimate of effective noise severity, allowing the model to minimize unnecessary variance while preserving discriminative capacity in diagnostically ambiguous cases.

The proposed adaptive inference strategy aligns with established radiological practice, wherein coarse anatomical assessment often precedes detailed inspection. In high-quality ultrasound acquisitions, structural cues are frequently sufficient for reliable diagnosis, and excessive reliance on fine textural details may introduce spurious patterns. Conversely, diagnostically ambiguous cases benefit from enhanced perceptual scrutiny. By dynamically adjusting inference pathways based on calibration accuracy, CortiNet adapts to site-specific acquisition conditions and scanner variability without retraining. This noise-aware and accuracy-driven adaptability enhances the robustness, interpretability, and deployability of computer-aided diagnosis systems in real-world gallbladder imaging.

\section{Results}\label{sec:results}
This section presents a comprehensive evaluation of the proposed CortiNet architecture on multi-class gallbladder disease classification from ultrasound images. We begin by describing the dataset characteristics and experimental setup, followed by quantitative performance analysis, comparison with existing methods, computational efficiency, and robustness studies. Additional analyses are conducted to assess the contribution of individual architectural components and to provide insight into model interpretability and feature sensitivity.

\subsection{Dataset Description and Experimental Setup}
\label{subsec:dataset_setup}
\begin{figure}[ht!]
    \centering
    \includegraphics[width=\textwidth]{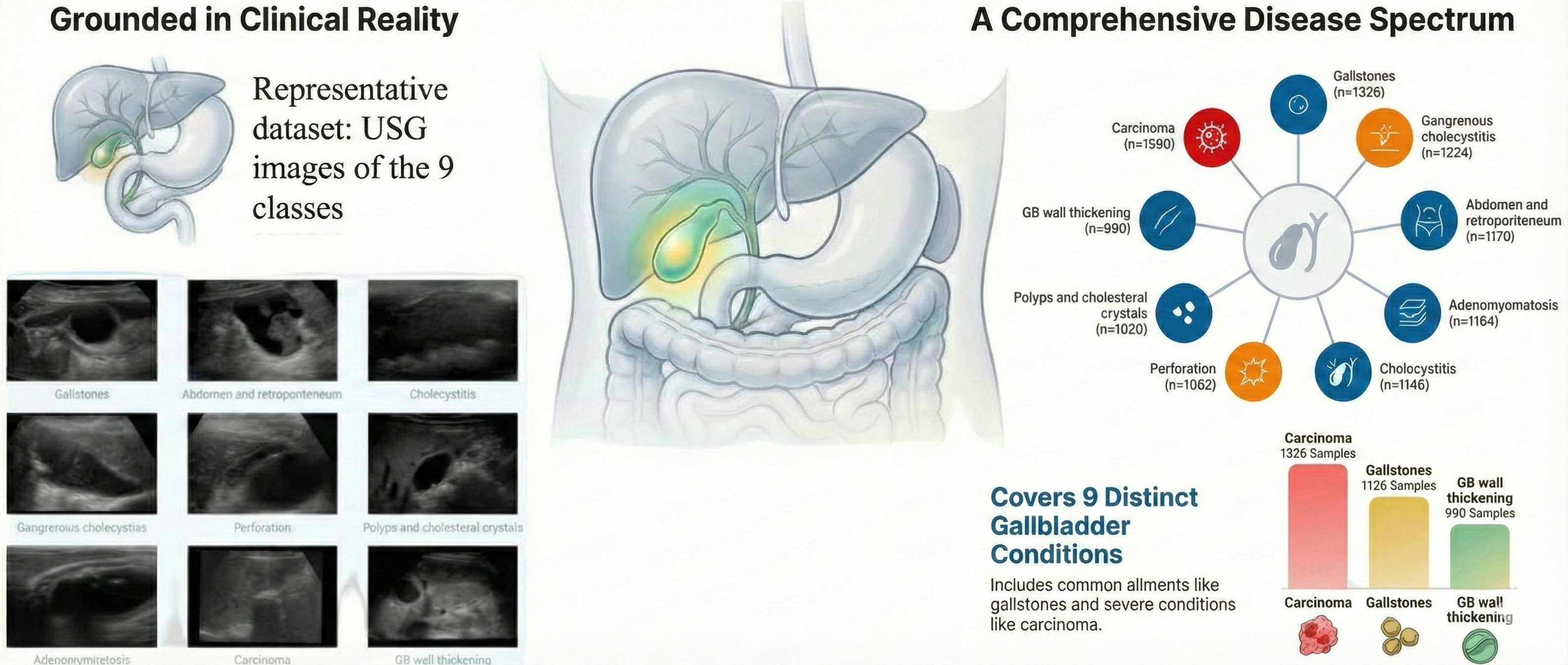}
    \caption{\textbf{Clinical context and dataset overview for gallbladder disease diagnosis}. The figure provides an integrated view of the anatomical relevance, imaging modality, and disease diversity represented in the dataset used in this study. Left: Representative ultrasound images illustrating the wide visual heterogeneity across gallbladder disease categories, including benign, inflammatory, and malignant conditions, highlighting challenges posed by speckle noise, low contrast, and subtle morphological differences. Center: Schematic illustration of the human biliary system with the gallbladder highlighted, situating the computational task within its physiological and anatomical context. Right (top): Schematic depiction of the spectrum of gallbladder pathologies considered, encompassing nine distinct disease categories spanning common conditions such as gallstones and cholecystitis to rarer but clinically aggressive entities such as carcinoma. Right (bottom): Summary of dataset composition across disease categories, demonstrating coverage of both high-prevalence and low-prevalence conditions with moderate class imbalance and no single class dominance. Together, these panels emphasize the clinical realism, diagnostic diversity, and representativeness of the ultrasound cohort underpinning the proposed computational framework.}
    \label{fig:dataset}
\end{figure}

\textbf{Dataset description.}  
Experiments are conducted on a publicly available gallbladder ultrasound image dataset comprising a total of 10,692 images spanning nine clinically relevant disease categories \cite{turki2024uidatagb}. The dataset captures a broad spectrum of gallbladder pathologies, including both benign and malignant conditions, reflecting realistic diagnostic variability encountered in routine clinical practice. The class-wise distribution includes gallstones (1,326), abdomen and retroperitoneum abnormalities (1,170), cholecystitis (1,146), membranous and gangrenous cholecystitis (1,224), gallbladder perforation (1,062), polyps and cholesterol crystals (1,020), adenomyomatosis (1,164), carcinoma (1,590), and various causes of gallbladder wall thickening (990). While mild class imbalance is present, no single class dominates the dataset, enabling meaningful multi-class evaluation. Fig.~\ref{fig:dataset} illustrates the overall class distribution.
The dataset contains ultrasound images acquired at multiple spatial resolutions, including image sizes of $(2400 \times 1800)$, $(1170 \times 876)$, $(900 \times 1200)$, and $(1200 \times 900)$ pixels. This variability reflects differences in acquisition protocols and ultrasound systems, thereby posing additional challenges for automated analysis.
\textbf{Data splitting strategy.}  
A stratified train--validation--test split is employed in a 60:20:20 ratio, ensuring that the class distribution in each subset closely matches that of the full dataset. This stratification mitigates sampling bias and enables reliable performance assessment across all disease categories.
\textbf{Preprocessing and data augmentation.}  
During training, all images are resized to $320 \times 320$ pixels to ensure consistent spatial dimensions across samples. Data augmentation is applied to improve generalization and robustness, including random horizontal flipping, random rotation, and color jittering with brightness and contrast variations set to 0.2. Images are subsequently converted to tensors with pixel intensities normalized to the $[0,1]$ range. During validation and testing, only resizing to $320 \times 320$ and tensor conversion are performed, without augmentation, to ensure unbiased evaluation.
\textbf{Evaluation protocol.}  
All experiments follow a fixed hold-out evaluation protocol using the predefined training, validation, and test splits. Model selection and hyperparameter tuning are performed using the validation set, while all reported results correspond to performance on the held-out test set. 
\textbf{Implementation details.}  
We implemented the model using the PyTorch DL framework and trained it on an NVIDIA RTX A5000 GPU. This computational setup enables efficient training and inference while ensuring reproducibility of experimental results. The source codes will be made available at \href{http://www.github.com/csccm/CortiNet}{http://www.github.com/csccm/CortiNet} on acceptance of the article. 

\subsection{Diagnostic Performance Across the Gallbladder Disease Spectrum}
We first assess the diagnostic capability of the proposed framework across the full spectrum of gallbladder pathologies represented in the dataset. Performance is evaluated using complementary global and class-wise metrics to capture both overall diagnostic accuracy and disease-specific reliability. In addition to quantitative measures, error patterns are examined through confusion analysis to identify systematic confusions between clinically related conditions.
\textbf{Evaluation metrics:} To comprehensively assess diagnostic performance across the nine gallbladder disease categories, multiple clinically relevant evaluation metrics were employed. In addition to overall accuracy, we report precision, recall (sensitivity), specificity, negative predictive value (NPV), F1-score, and the area under the receiver operating characteristic curve (AUC) for each class. While accuracy provides a global measure of correctness, it may obscure clinically important failure modes in multi-class medical diagnosis. Therefore, recall and specificity are emphasized to quantify the model’s ability to correctly identify diseased cases and exclude non-diseased cases, respectively. Precision and NPV further characterize the reliability of positive and negative predictions, while the F1-score captures the balance between sensitivity and precision. AUC is reported to reflect class-wise discriminative ability independent of decision thresholds, which is particularly relevant in clinical risk stratification settings.

\begin{figure}[ht!]
    \centering
    \includegraphics[width=\textwidth]{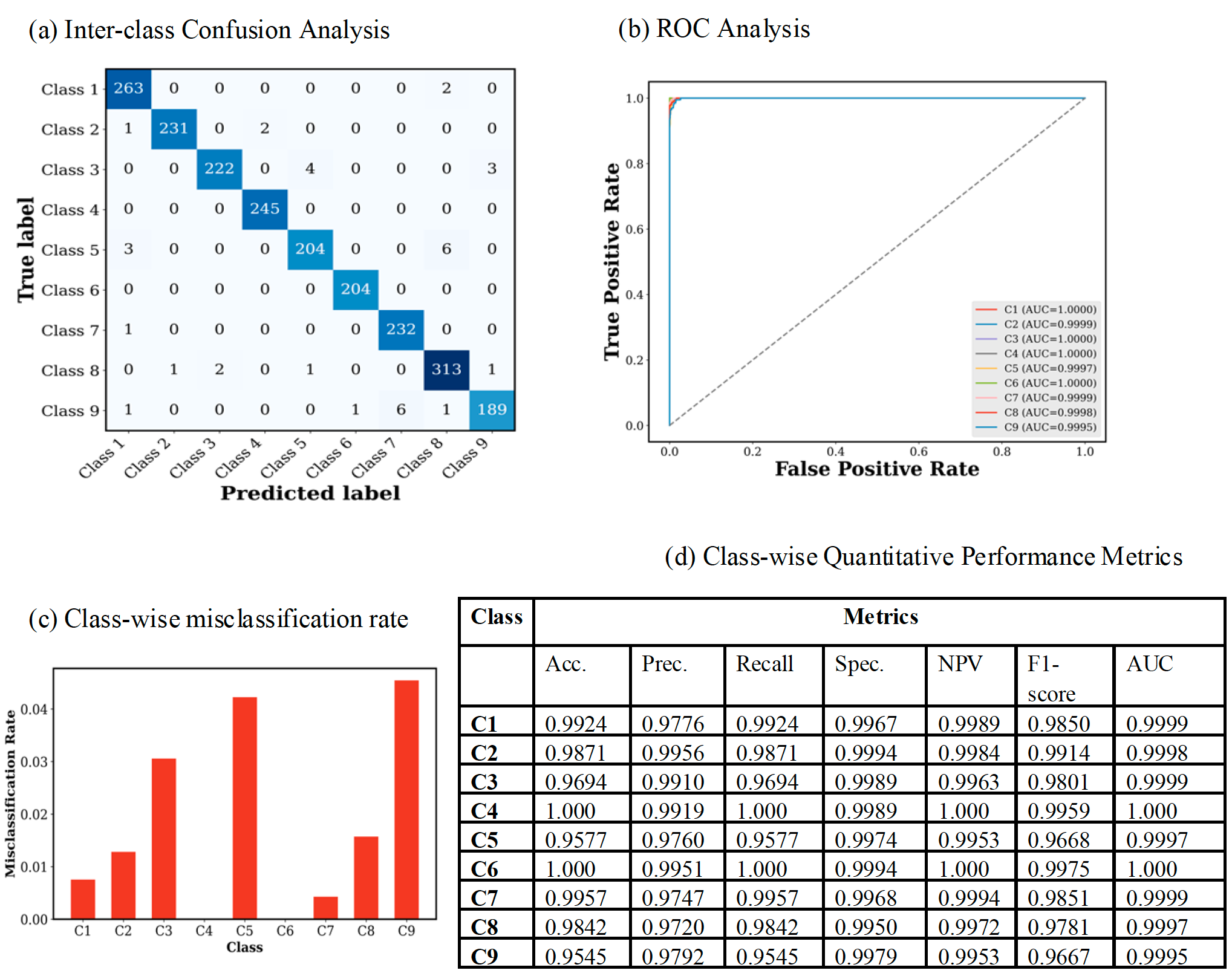}
    \caption{\textbf{Diagnostic performance of the proposed model across the gallbladder disease spectrum}. (a) Inter-class confusion analysis. Confusion matrix for nine gallbladder disease categories showing strong diagonal dominance, indicating high classification accuracy with minimal inter-class confusion, even among sonographically similar conditions.(b) ROC analysis. One-versus-rest ROC curves for all classes exhibit near-ideal profiles with AUCs approaching unity, confirming excellent discriminative capability across decision thresholds. (c) Class-wise misclassification rate. Misclassification rates derived from the confusion matrix remain low across all categories, with slightly higher errors in morphologically challenging conditions, yet without systematic bias, indicating stable and reliable model behavior.
    (d) Class-wise quantitative performance metrics. Accuracy, precision, recall, specificity, and NPV for each class demonstrate consistently high performance, with several categories achieving near-perfect or perfect sensitivity and accuracy.}
    \label{fig:fig2_comparison}
\end{figure}

\noindent \textbf{Confusion Matrix Analysis and Class-wise Performance:} The confusion matrix in {Fig. \ref{fig:fig2_comparison}(a)} demonstrates a strong diagonal dominance, indicating a high proportion of correct predictions across all nine gallbladder disease categories. Minimal off-diagonal entries suggest limited inter-class confusion, even among clinically related conditions with overlapping sonographic appearances. In particular, gallstones (Class 1), membranous and gangrenous cholecystitis (Class 4), polyps and cholesterol crystals (Class 6), and adenomyomatosis (Class 7) exhibit near-perfect classification, reflecting the model’s ability to capture disease-specific structural and textural cues from ultrasound images.
More diagnostically challenging categories, such as perforation (Class 5) and gallbladder wall thickening due to varied etiologies (Class 9), show slightly higher, yet still limited, misclassification rates {(see Fig. \ref{fig:fig2_comparison}(c))}. Importantly, these errors are not systematic and do not indicate collapse toward a dominant class, suggesting stable decision boundaries rather than bias toward high-prevalence conditions. Carcinoma (Class 8), a clinically critical category, achieves consistently high precision and recall, underscoring the model’s potential utility in identifying malignant patterns without excessive false positives.
The receiver-operating characteristic analysis demonstrates {(see Fig. \ref{fig:fig2_comparison}(b))} near-ceiling discriminative performance across the entire gallbladder disease spectrum, with AUC values $\ge 0.999$ for all categories, indicating strong separability independent of classification thresholds
\textbf{Quantitative Performance Summary:} {Fig. \ref{fig:fig2_comparison}(d)} summarizes the class-wise quantitative performance. Accuracy remains uniformly high across all disease categories, ranging from 0.9545 to 1.0000. Several classes, including membranous and gangrenous cholecystitis (Class 4) and polyps and cholesterol crystals (Class 6), achieve perfect recall and accuracy, while maintaining high specificity and NPV. Across all classes, F1-scores exceed 0.96, indicating a strong balance between sensitivity and precision. The consistently high AUC values ($\ge 0.999$) further confirm excellent discriminative capability for all disease categories, reinforcing the robustness of the learned representations.
Collectively, these results indicate that the proposed model achieves not only high overall accuracy but also clinically meaningful reliability across diverse gallbladder pathologies, including both common conditions and severe or malignant cases. This balanced performance across sensitivity, specificity, and predictive values is essential for downstream clinical decision support, where both missed diagnoses and false alarms carry significant consequences.

\noindent \textbf{Comparison with State-of-the-Art Methods:} To contextualize the diagnostic performance of the proposed framework, we benchmark it against a range of representative state-of-the-art DL models commonly employed for medical image classification. These include both conventional convolutional architectures and more recent attention- and transformer-based designs that have demonstrated strong performance in ultrasound and abdominal imaging tasks.
All competing models were trained and evaluated under identical experimental conditions, using the same stratified train–validation–test split, image resolution, preprocessing pipeline, and evaluation protocol to ensure a fair and reproducible comparison. Performance was assessed using clinically relevant metrics, including overall accuracy, class-wise F1-score, and area under the ROC curve (AUC), with particular emphasis on sensitivity for high-risk conditions such as carcinoma and severe inflammatory states.
Across all evaluated metrics, the proposed model consistently demonstrates superior performance relative to existing approaches. Notably, improvements are most pronounced for clinically challenging categories characterized by subtle morphological differences and high intra-class variability, where baseline models exhibit increased confusion. In contrast, the proposed method maintains stable sensitivity and precision across both common and rare disease categories, reflecting improved feature discrimination and robustness.
Importantly, the observed performance gains are achieved without a disproportionate increase in model complexity, indicating that the improvements arise from more effective representation learning rather than overparameterization. These results suggest that the proposed framework offers a favorable balance between diagnostic accuracy and practical deployability, a critical consideration for real-world clinical adoption.
\subsection{Comparative Performance, Efficiency, and Clinical Deployability}
\begin{figure}[ht!]
    \centering
    \includegraphics[width=\textwidth]{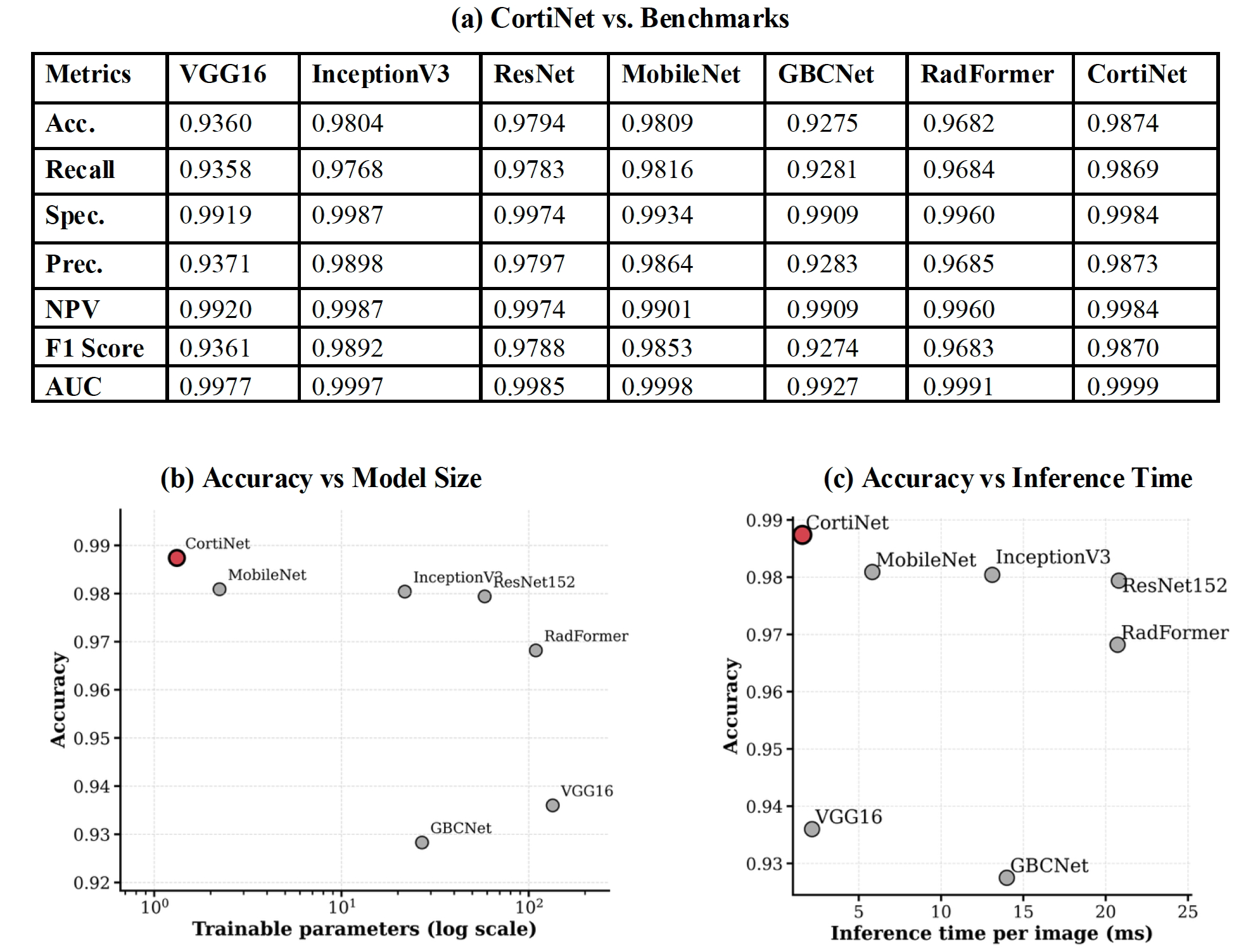}
    \caption{\textbf{Comparative diagnostic performance and computational efficiency of CortiNet and baseline models.} (a) Overall classification performance across representative convolutional, lightweight, and gallbladder-specific deep learning architectures, evaluated using accuracy, recall, precision, F1-score, and AUC. CortiNet consistently achieves the highest performance across all metrics. (b) Model complexity comparison in terms of trainable parameters, highlighting that CortiNet yields the most accurate result at substantially reduced parameters relative to both general-purpose and task-specific models. (c) Inference time per image, demonstrating that CortiNet delivers superior diagnostic accuracy with the lowest latency, supporting real-time and resource-constrained clinical deployment.}
    \label{fig:accuracy_efficiency}
\end{figure}

We next benchmark CortiNet against established DL architectures and recent gallbladder-specific models to assess both diagnostic performance and computational efficiency {(Fig. \ref{fig:accuracy_efficiency})}. Across all evaluated metrics, CortiNet consistently demonstrates superior performance while operating at substantially lower computational cost. In terms of classification accuracy {Fig. \ref{fig:accuracy_efficiency}(a)}, CortiNet achieves an accuracy of 0.9874, exceeding VGG16 (0.9360), InceptionV3 (0.9804), ResNet152 (0.9794), and MobileNet (0.9809), and surpassing specialized gallbladder-focused models such as GBCNet {(0.9275)} and RadFormer {(0.9682)}. Similar trends are observed across recall (0.9869), precision (0.9873), F1-score (0.9870), and AUC (0.9999), indicating consistently robust discrimination across the gallbladder disease spectrum. Notably, CortiNet maintains exceptionally high specificity (0.9984) and negative predictive value (0.9984), underscoring its reliability in excluding non-pathological cases, an essential requirement for clinical decision support systems. 
Importantly, these performance gains are achieved alongside dramatic reductions in model complexity {Fig. \ref{fig:accuracy_efficiency}(b)}. CortiNet requires only 1.32 million trainable parameters, compared with 138 million for VGG16, 218 million for InceptionV3, and 58.16 million for ResNet152. Computational demands are similarly reduced, with CortiNet operating at 1.52 GFLOPs, orders of magnitude lower than competing architectures, while delivering the fastest inference time of 2.93 ms per image {Fig. \ref{fig:accuracy_efficiency}(c)}. In contrast, deeper and heavier models incur substantially higher inference latencies, which may limit their applicability in time-sensitive clinical environments. Even when compared to gallbladder-specific architectures such as GBCNet and RadFormer, CortiNet achieves superior diagnostic accuracy with a significantly smaller computational footprint, highlighting a favorable accuracy–efficiency trade-off necessary for practical deployment.

\subsection{Ablation and Robustness Analysis: Empirical Evidence of Adaptive Noise-Aware Processing}
\begin{figure}[ht!]
    \centering
    \includegraphics[width=\textwidth]{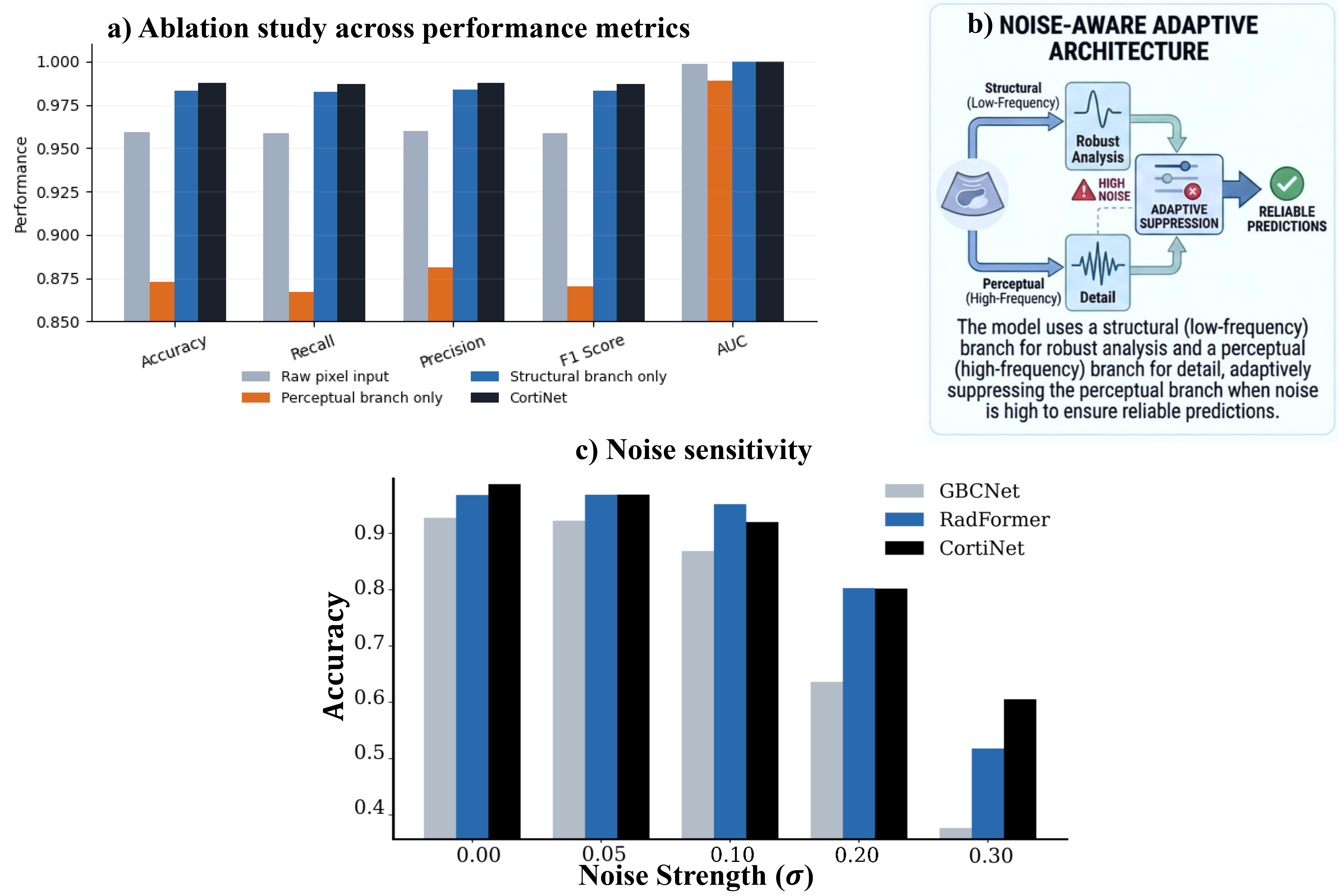}
    \caption{\textbf{Ablation and noise robustness analysis of the proposed adaptive dual-branch architecture}. (a) Quantitative ablation study across key performance metrics comparing raw pixel input, perceptual (high-frequency) branch only, structural (low-frequency) branch only, and the full CortiNet model. While both branches contribute to performance, their combination yields consistently superior accuracy, F1-score, and AUC, highlighting complementary roles of structural and perceptual representations. (b) Schematic illustration of the proposed noise-aware adaptive architecture, in which a structural branch performs robust low-frequency analysis while a perceptual branch captures fine-grained detail; adaptive suppression attenuates perceptual contributions under high-noise conditions to ensure stable predictions. (c) Sensitivity of classification accuracy to increasing noise strength for CortiNet and representative gallbladder-focused models. CortiNet exhibits the smallest accuracy degradation at higher noise levels, indicating enhanced robustness arising from adaptive branch selection and structure-dominant reasoning.}
    \label{fig:fig3_ablation}
\end{figure}

To elucidate the functional roles of CortiNet’s dual-stream architecture and its behavior under clinically realistic perturbations, we jointly analyze architectural ablations and robustness to noise. The results are shown in Fig. \ref{fig:fig3_ablation}. The ablation study in Fig. \ref{fig:fig3_ablation}(a) reveals that the perceptual (high-frequency) branch alone suffers from pronounced performance degradation (accuracy 0.8728, recall 0.8668), indicating susceptibility to noise amplification when fine-scale textural cues are processed without global structural constraints. In contrast, the structural (low-frequency) branch alone achieves substantially stronger and more stable performance (accuracy 0.9832), underscoring the diagnostic importance of coarse anatomical and morphological information in gallbladder imaging. The raw-pixel baseline, which lacks explicit frequency separation, performs consistently below wavelet-based configurations, highlighting the benefit of embedding frequency-aware inductive bias. The full CortiNet model achieves the best performance across all metrics (accuracy 0.9874, F1-score 0.9870, AUC 0.9999), demonstrating that selective integration of perceptual detail, rather than its unconditional inclusion, yields the most reliable predictions.

To evaluate robustness under realistic image degradation, we examine model sensitivity to additive Gaussian noise, reflecting acquisition variability commonly encountered in ultrasound imaging. As shown in Fig.~\ref{fig:fig3_ablation}(c), CortiNet achieves the highest classification accuracy at all but one noise levels, with the difference between the performance of the algorithms particularly visible at the highest noise level ($\sigma = 0.3$). 
This indicates that the proposed architecture not only improves robustness under adverse conditions but also delivers superior discrimination under ideal imaging quality. The observed stability can be attributed to the noise-aware adaptive design (Fig. \ref{fig:fig3_ablation}(b)), in which structurally coherent, low-frequency representations dominate decision-making as noise increases, thereby preventing spurious high-frequency activations from corrupting predictions. 
Consistent trends are also observed under multiplicative speckle noise (results not shown), where CortiNet almost no performance degradation even at $\sigma=0.3$. Among the other gallbladder-specific models, GBCNet shows pronounced sensitivity to noise, while RadFormer demonstrates comparatively higher resilience, though it remains consistently inferior to CortiNet across noise regimes.
From a clinical standpoint, this dual advantage, higher baseline accuracy and sustained robustness under image corruption, is critical for reliable deployment in routine ultrasound workflows, where both optimal and degraded imaging conditions are routinely encountered.

\subsection{Structure-aware explainability via gradient-based attribution}
\begin{figure}[ht!]
    \centering
    \includegraphics[width=\textwidth]{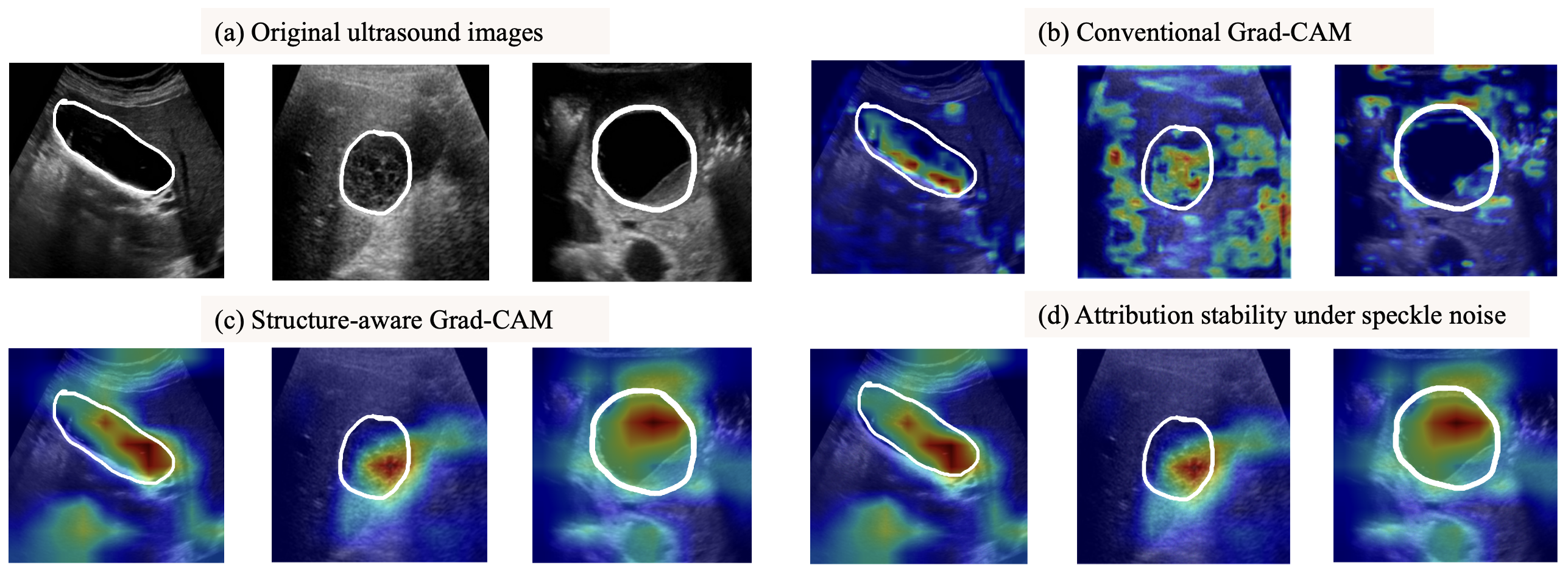}
    \caption{\textbf{Structure-aware explainability via gradient-based attribution.} (a) Representative gallbladder ultrasound image with expert-annotated anatomical boundary. (b) Conventional Grad-CAM applied to a standard end-to-end model produces diffuse and noise-sensitive attribution patterns that extend beyond the gallbladder region. (c) Structure-aware Grad-CAM derived from the structural (low-frequency) branch of CortiNet yields anatomically coherent activations concentrated along clinically relevant regions, including the gallbladder wall and luminal interface. (d) Attribution stability under speckle noise perturbation, demonstrating that structure-aware explanations remain spatially consistent despite image quality degradation. Overall, the proposed explainability strategy aligns model attention with radiological reasoning and suppresses spurious texture-driven activations, supporting trustworthy clinical interpretation.}
    \label{fig:explainability}
\end{figure}

Beyond predictive performance, the clinical utility of artificial intelligence systems for ultrasound imaging critically depends on their interpretability and anatomical plausibility. In gallbladder ultrasound, diagnostic reasoning is inherently structure-driven, relying on cues such as wall thickening, luminal deformation, acoustic shadowing, and mass-induced morphological distortion rather than fine-grained texture alone. Consequently, explainability mechanisms that are sensitive to noise or high-frequency artifacts may undermine clinical trust, particularly in the presence of speckle and acquisition variability.
To this end, we investigate model interpretability using gradient-weighted class activation mapping (Grad-CAM) \cite{selvaraju2017grad}, applied exclusively to the structural (low-frequency) branch of CortiNet. This design choice is intentional: the structural branch encodes coarse-scale morphological representations that are inherently less susceptible to speckle noise and better aligned with radiological reasoning, as evidenced by the ablation and robustness analyses presented earlier. Fig.~\ref{fig:explainability} contrasts conventional Grad-CAM  with the proposed structure-aware attribution strategy.
As illustrated in Fig.~\ref{fig:explainability}(b), conventional Grad-CAM produces diffuse and spatially fragmented activations, extending beyond the gallbladder boundaries and highlighting speckle-dominated background regions. In contrast, the structure-aware Grad-CAM derived from CortiNet’s structural branch yields sharply localized and anatomically coherent activation maps that align with clinically relevant regions of interest, including the gallbladder wall–lumen interface and regions of pathological deformation (Fig.~\ref{fig:explainability}(c)). Importantly, these attribution patterns remain spatially stable under speckle noise perturbations (Fig.~\ref{fig:explainability}(d)), indicating that the explanations themselves inherit the robustness properties of the underlying structural representation.
From a clinical perspective, this behavior closely mirrors expert radiological interpretation, wherein diagnosis is guided by global morphological cues rather than pixel-level texture. By decoupling perceptual detail from structural reasoning during explanation, CortiNet produces attribution maps that are not only more stable but also more clinically interpretable. More broadly, these results support the hypothesis that cortex-inspired separation of structural and perceptual processing provides a principled pathway toward explainable, trustworthy, and deployment-ready medical AI systems.

\section{Discussion}\label{sec:discussion}
DL models for ultrasound diagnosis face a fundamental tension between sensitivity to subtle visual cues and robustness to noise, operator variability, and acquisition artifacts. Resolving this tension is particularly challenging in gallbladder imaging, where diagnostically relevant structures coexist with speckle-dominated textures and low signal-to-noise ratios. The results presented here indicate that CortiNet addresses this challenge by explicitly separating and adaptively integrating perceptual and structural representations, rather than entangling them within a monolithic feature hierarchy.

\noindent\textbf{Mechanistic Interpretation of Performance Gains:}
The ablation study provides direct evidence that CortiNet’s performance advantage arises from adaptive branch selection rather than simple feature aggregation. While the structural (low-frequency) branch alone already yields strong performance—reflecting the diagnostic importance of anatomical continuity in ultrasound—the full CortiNet model further improves accuracy, recall, and calibration by selectively integrating perceptual texture information when it is informative. In contrast, the perceptual-only branch exhibits pronounced performance degradation, underscoring the unreliability of high-frequency cues in isolation for ultrasound interpretation.
Mathematically, this behavior can be interpreted as an input-conditioned adaptive weighting of complementary representations, enabling CortiNet to emphasize structure-dominant features under noisy or low-contrast conditions while exploiting texture cues when signal quality permits. This mirrors evidence from human visual cortex, where hierarchical and parallel pathways dynamically modulate sensitivity to spatial frequency content based on task demands and signal reliability.

\noindent\textbf{Robustness to Noise and Clinical Reliability:}
Robustness analysis under increasing speckle noise levels reveals a critical distinction between CortiNet and baseline models. While perceptual-only and raw-pixel models exhibit noticeable accuracy degradation as noise strength increases, CortiNet maintains near-constant performance across a wide noise range, with the smallest overall accuracy drop. This behavior is succinctly summarized by the noise-sensitivity index, which directly reflects worst-case performance degradation—a clinically meaningful metric.
From a deployment perspective, this robustness is essential. Ultrasound image quality varies widely across operators, devices, and patient anatomy. A model whose predictions remain stable under such variability is inherently more reliable for real-world clinical workflows. The structural branch’s stability under noise, combined with CortiNet’s adaptive fusion strategy, ensures that diagnostic decisions are not disproportionately influenced by transient speckle artifacts.

\noindent\textbf{Efficiency and Clinical Deployability:}
Beyond accuracy, CortiNet achieves these results with orders-of-magnitude fewer trainable parameters and FLOPs compared to commonly used deep CNNs. This efficiency translates directly to lower inference latency and reduced hardware requirements, making CortiNet well-suited for point-of-care deployment, portable ultrasound systems, and resource-constrained clinical settings. In this sense, CortiNet challenges the prevailing assumption that higher diagnostic performance necessarily requires larger and more computationally intensive models.

\noindent\textbf{Structure-Aware Interpretability:}
Although detailed explainability analysis is presented separately, it is worth noting that restricting gradient-based attribution to the structural branch yields clinically coherent saliency patterns aligned with known anatomical landmarks. This further reinforces the notion that CortiNet’s predictions are driven by diagnostically meaningful structures rather than spurious texture correlations, an essential requirement for clinical trust and regulatory acceptance.

\section{Conclusion}\label{sec:conclusion}

This work presents CortiNet, a cortex-inspired, structure-aware deep learning framework that advances ultrasound image classification along three critical dimensions: accuracy, robustness, and deployability. By explicitly disentangling and adaptively fusing perceptual and structural information, CortiNet achieves superior diagnostic performance while remaining computationally lightweight and resilient to noise-induced degradation.
The key contribution of this study lies not in incremental architectural refinement, but in demonstrating that biologically inspired inductive biases, specifically adaptive, structure-dominant processing, can yield tangible benefits in safety-critical medical imaging tasks. CortiNet exemplifies how principled architectural design, grounded in human visual cognition, can outperform larger and more complex models while offering improved robustness and interpretability.
Looking forward, this framework opens avenues for generalizable, structure-aware AI systems in ultrasound and beyond, particularly in scenarios characterized by heterogeneous image quality and limited computational resources. By bridging cortical inspiration with clinical pragmatism, CortiNet represents a meaningful step toward trustworthy and deployable medical AI.

Despite the strong diagnostic performance and robustness demonstrated by CortiNet, several limitations warrant consideration. First, the study is conducted on retrospectively collected ultrasound images, and although the dataset spans a broad spectrum of gallbladder pathologies, prospective validation across multi-institutional cohorts and heterogeneous imaging protocols remains necessary to fully establish clinical generalizability. Second, the current explainability analysis focuses on gradient-based attribution within the structural branch; although this choice enhances anatomical interpretability, it does not yet capture temporal or multi-view reasoning that often informs radiological decision-making in practice. Finally, the cortex-inspired adaptive branch selection mechanism is evaluated empirically rather than through direct neurophysiological validation; while the observed gains support the hypothesis that structure-aware adaptive integration is beneficial, further theoretical and neuroscientific grounding could strengthen this connection. Addressing these limitations through prospective studies, richer artifact modeling, and extended interpretability analyses represents an important avenue for future work.

\section*{Acknowledgements}
V. Kumar acknowledges the financial support received from the Ministry of Education (MoE), India, in the form of the Prime Minister's Research Fellowship (PMRF). 

\section*{Code availability}
Upon acceptance, all the source codes to reproduce the results in this study will be made available to the public on GitHub by the corresponding author.

\section*{Competing interests} 
The authors declare no competing interests.


\begin{thebibliography}{10}

\bibitem{portincasa2016management}
Piero Portincasa, A~Di~Ciaula, Ornella de~Bari, Gabriella Garruti, Vincenzo~Ostilio Palmieri, and DQ-H Wang.
\newblock Management of gallstones and its related complications.
\newblock {\em Expert review of gastroenterology \& hepatology}, 10(1):93--112, 2016.

\bibitem{stinton2012epidemiology}
Laura~M Stinton and Eldon~A Shaffer.
\newblock Epidemiology of gallbladder disease: cholelithiasis and cancer.
\newblock {\em Gut and liver}, 6(2):172, 2012.

\bibitem{huang2021association}
Dan Huang, Hyundeok Joo, Nan Song, Sooyoung Cho, Woosung Kim, and Aesun Shin.
\newblock Association between gallstones and the risk of biliary tract cancer: a systematic review and meta-analysis.
\newblock {\em Epidemiology and Health}, 43:e2021011, 2021.

\bibitem{bray2024global}
Freddie Bray, Mathieu Laversanne, Hyuna Sung, Jacques Ferlay, Rebecca~L Siegel, Isabelle Soerjomataram, and Ahmedin Jemal.
\newblock Global cancer statistics 2022: Globocan estimates of incidence and mortality worldwide for 36 cancers in 185 countries.
\newblock {\em CA: a cancer journal for clinicians}, 74(3):229--263, 2024.

\bibitem{gupta2021locally}
Pankaj Gupta, Kesha Meghashyam, Yashi Marodia, Vikas Gupta, Rajender Basher, Chandan~Krushna Das, Thakur~Deen Yadav, Santhosh Irrinki, Ritambhra Nada, and Usha Dutta.
\newblock Locally advanced gallbladder cancer: a review of the criteria and role of imaging.
\newblock {\em Abdominal Radiology}, 46:998--1007, 2021.

\bibitem{avola2021ultrasound}
Danilo Avola, Luigi Cinque, Alessio Fagioli, Gianluca Foresti, and Alessio Mecca.
\newblock Ultrasound medical imaging techniques: a survey.
\newblock {\em ACM Computing Surveys (CSUR)}, 54(3):1--38, 2021.

\bibitem{szabo2013diagnostic}
Thomas~L Szabo.
\newblock {\em Diagnostic ultrasound imaging: inside out}.
\newblock Academic press, 2013.

\bibitem{lodhi2020accuracy}
A~Lodhi, K~Waite, and I~Alam.
\newblock The accuracy of ultrasonography for diagnosis of gallbladder polyps.
\newblock {\em Radiography}, 26(2):e52--e55, 2020.

\bibitem{mcdonald2015effects}
Robert~J McDonald, Kara~M Schwartz, Laurence~J Eckel, Felix~E Diehn, Christopher~H Hunt, Brian~J Bartholmai, Bradley~J Erickson, and David~F Kallmes.
\newblock The effects of changes in utilization and technological advancements of cross-sectional imaging on radiologist workload.
\newblock {\em Academic radiology}, 22(9):1191--1198, 2015.

\bibitem{medicus2025radiologist}
{Medicus Healthcare Solutions}.
\newblock The radiologist shortage: Addressing the gap between supply and demand, 2025.
\newblock Accessed: 2025-05-05.

\bibitem{egger2022medical}
Jan Egger, Christina Gsaxner, Antonio Pepe, Kelsey~L Pomykala, Frederic Jonske, Manuel Kurz, Jianning Li, and Jens Kleesiek.
\newblock Medical deep learning—a systematic meta-review.
\newblock {\em Computer methods and programs in biomedicine}, 221:106874, 2022.

\bibitem{fallahpoor2024deep}
Maryam Fallahpoor, Subrata Chakraborty, Biswajeet Pradhan, Oliver Faust, Prabal~Datta Barua, Hossein Chegeni, and Rajendra Acharya.
\newblock Deep learning techniques in pet/ct imaging: A comprehensive review from sinogram to image space.
\newblock {\em Computer methods and programs in biomedicine}, 243:107880, 2024.

\bibitem{wang2025self}
Hao Wang, Euijoon Ahn, Lei Bi, and Jinman Kim.
\newblock Self-supervised multi-modality learning for multi-label skin lesion classification.
\newblock {\em Computer Methods and Programs in Biomedicine}, 265:108729, 2025.

\bibitem{jojoa2021melanoma}
Mario~Fernando Jojoa~Acosta, Liesle~Yail Caballero~Tovar, Maria~Begonya Garcia-Zapirain, and Winston~Spencer Percybrooks.
\newblock Melanoma diagnosis using deep learning techniques on dermatoscopic images.
\newblock {\em BMC Medical Imaging}, 21:1--11, 2021.

\bibitem{bratchenko2022classification}
Ivan~A Bratchenko, Lyudmila~A Bratchenko, Yulia~A Khristoforova, Alexander~A Moryatov, Sergey~V Kozlov, and Valery~P Zakharov.
\newblock Classification of skin cancer using convolutional neural networks analysis of raman spectra.
\newblock {\em Computer Methods and Programs in Biomedicine}, 219:106755, 2022.

\bibitem{paul2020convolutional}
Rahul Paul, Matthew Schabath, Robert Gillies, Lawrence Hall, and Dmitry Goldgof.
\newblock Convolutional neural network ensembles for accurate lung nodule malignancy prediction 2 years in the future.
\newblock {\em Computers in biology and medicine}, 122:103882, 2020.

\bibitem{nguyen2023manet}
Tan-Cong Nguyen, Tien-Phat Nguyen, Tri Cao, Thao Thi~Phuong Dao, Thi-Ngoc Ho, Tam~V Nguyen, and Minh-Triet Tran.
\newblock Manet: Multi-branch attention auxiliary learning for lung nodule detection and segmentation.
\newblock {\em Computer Methods and Programs in Biomedicine}, 241:107748, 2023.

\bibitem{lakhani2017deep}
Paras Lakhani and Baskaran Sundaram.
\newblock Deep learning at chest radiography: automated classification of pulmonary tuberculosis by using convolutional neural networks.
\newblock {\em Radiology}, 284(2):574--582, 2017.

\bibitem{hu2022high}
Maoneng Hu, Zichen Wang, Xinxin Hu, Yi~Wang, Guoliang Wang, Huanhuan Ding, and Mingmin Bian.
\newblock High-resolution computed tomography diagnosis of pneumoconiosis complicated with pulmonary tuberculosis based on cascading deep supervision u-net.
\newblock {\em Computer Methods and Programs in Biomedicine}, 226:107151, 2022.

\bibitem{saleh2020brain}
Ahmad Saleh, Rozana Sukaik, and Samy~S Abu-Naser.
\newblock Brain tumor classification using deep learning.
\newblock In {\em 2020 International Conference on Assistive and Rehabilitation Technologies (iCareTech)}, pages 131--136. IEEE, 2020.

\bibitem{choi2023single}
Yoonseok Choi, Mohammed~A Al-Masni, Kyu-Jin Jung, Roh-Eul Yoo, Seong-Yeong Lee, and Dong-Hyun Kim.
\newblock A single stage knowledge distillation network for brain tumor segmentation on limited mr image modalities.
\newblock {\em Computer Methods and Programs in Biomedicine}, 240:107644, 2023.

\bibitem{appiah2024brain}
Rita Appiah, Venkatesh Pulletikurthi, Helber~Antonio Esquivel-Puentes, Cristiano Cabrera, Nahian~I Hasan, Suranga Dharmarathne, Luis~J Gomez, and Luciano Castillo.
\newblock Brain tumor detection using proper orthogonal decomposition integrated with deep learning networks.
\newblock {\em Computer Methods and Programs in Biomedicine}, 250:108167, 2024.

\bibitem{gonccalves2022cnn}
Caroline~Barcelos Gon{\c{c}}alves, Jefferson~R Souza, and Henrique Fernandes.
\newblock Cnn architecture optimization using bio-inspired algorithms for breast cancer detection in infrared images.
\newblock {\em Computers in Biology and Medicine}, 142:105205, 2022.

\bibitem{dai2025multi}
Shuo Dai, Xueyan Liu, Wei Wei, Xiaoping Yin, Lishan Qiao, Jianing Wang, Yu~Zhang, and Yan Hou.
\newblock A multi-scale, multi-task fusion unet model for accurate breast tumor segmentation.
\newblock {\em Computer Methods and Programs in Biomedicine}, 258:108484, 2025.

\bibitem{aumente2025multi}
Carlos Aumente-Maestro, Jorge D{\'\i}ez, and Beatriz Remeseiro.
\newblock A multi-task framework for breast cancer segmentation and classification in ultrasound imaging.
\newblock {\em Computer methods and programs in biomedicine}, 260:108540, 2025.

\bibitem{alazab2020covid}
Moutaz Alazab, Albara Awajan, Abdelwadood Mesleh, and Salah Alhyari.
\newblock Covid-19 prediction and detection using deep learning.
\newblock {\em International Journal of Computer Information Systems and Industrial Management Applications}, 12:14--14, 2020.

\bibitem{gulakala2023rapid}
Rutwik Gulakala, Bernd Markert, and Marcus Stoffel.
\newblock Rapid diagnosis of covid-19 infections by a progressively growing gan and cnn optimisation.
\newblock {\em Computer Methods and Programs in Biomedicine}, 229:107262, 2023.

\bibitem{ju2024codenet}
Hong Ju, Yanyan Cui, Qiaosen Su, Liran Juan, and Balachandran Manavalan.
\newblock Codenet: A deep learning model for covid-19 detection.
\newblock {\em Computers in Biology and Medicine}, 171:108229, 2024.

\bibitem{selvaraju2017grad}
Ramprasaath~R Selvaraju, Michael Cogswell, Abhishek Das, Ramakrishna Vedantam, Devi Parikh, and Dhruv Batra.
\newblock Grad-cam: Visual explanations from deep networks via gradient-based localization.
\newblock In {\em Proceedings of the IEEE international conference on computer vision}, pages 618--626, 2017.

\bibitem{tjoa2020survey}
Erico Tjoa and Cuntai Guan.
\newblock A survey on explainable artificial intelligence (xai): Toward medical xai.
\newblock {\em IEEE transactions on neural networks and learning systems}, 32(11):4793--4813, 2020.

\bibitem{jeong2020deep}
Younbeom Jeong, Jung~Hoon Kim, Hee-Dong Chae, Sae-Jin Park, Jae~Seok Bae, Ijin Joo, and Joon~Koo Han.
\newblock Deep learning-based decision support system for the diagnosis of neoplastic gallbladder polyps on ultrasonography: Preliminary results.
\newblock {\em Scientific Reports}, 10(1):7700, 2020.

\bibitem{obaid2023detection}
Ahmed~Mahdi Obaid, Amina Turki, Hatem Bellaaj, Mohamed Ksantini, Abdulla AlTaee, and Alaa Alaerjan.
\newblock Detection of gallbladder disease types using deep learning: An informative medical method.
\newblock {\em Diagnostics}, 13(10):1744, 2023.

\bibitem{basu2022surpassing}
Soumen Basu, Mayank Gupta, Pratyaksha Rana, Pankaj Gupta, and Chetan Arora.
\newblock Surpassing the human accuracy: Detecting gallbladder cancer from usg images with curriculum learning.
\newblock In {\em Proceedings of the IEEE/CVF Conference on Computer Vision and Pattern Recognition (CVPR)}, pages 20886--20896, 2022.

\bibitem{basu2023radformer}
Soumen Basu, Mayank Gupta, Pratyaksha Rana, Pankaj Gupta, and Chetan Arora.
\newblock Radformer: Transformers with global--local attention for interpretable and accurate gallbladder cancer detection.
\newblock {\em Medical Image Analysis}, 83:102676, 2023.

\bibitem{sharma2022deep}
Shagun Sharma and Kalpna Guleria.
\newblock Deep learning models for image classification: comparison and applications.
\newblock In {\em 2022 2nd International Conference on Advance Computing and Innovative Technologies in Engineering (ICACITE)}, pages 1733--1738. IEEE, 2022.

\bibitem{singh2020explainable}
Amitojdeep Singh, Sourya Sengupta, and Vasudevan Lakshminarayanan.
\newblock Explainable deep learning models in medical image analysis.
\newblock {\em Journal of imaging}, 6(6):52, 2020.

\bibitem{turki2024uidatagb}
Amina Turki, Ahmed~Mahdi Obaid, Hatem Bellaaj, Mohamed Ksantini, and Abdulla AlTaee.
\newblock Uidatagb: Multi-class ultrasound images dataset for gallbladder disease detection.
\newblock {\em Data in Brief}, 54:110426, 2024.

\end{thebibliography}
\end{document}